\documentclass[11pt,a4paper]{article}
\usepackage[hyperref]{acl2017}
\usepackage{times}
\usepackage{latexsym}
\usepackage{amsmath, amsfonts, amssymb}
\usepackage{booktabs, multirow, tabularx, makecell}
\usepackage{siunitx}
\usepackage{graphicx}
\usepackage[labelfont=bf]{caption}
\usepackage{url}
\DeclareCaptionFont{ninept}{\fontsize{9pt}{11pt}\selectfont #1}
\newcommand{\R}{\mathbb{R}}

\aclfinalcopy 

\title{A Deep Neural Network Approach To Parallel Sentence Extraction}

\author{Francis Gr\'egoire \and Philippe Langlais \\
  Universit\'e de Montr\'eal \\
  {\tt \{gregoifr, felipe\}@iro.umontreal.ca}}

\date{}

\begin{document}

\maketitle

\begin{abstract}
\fontsize{10pt}{12pt}\selectfont
Parallel sentence extraction is a task addressing the data sparsity problem found in multilingual natural language processing applications. We propose an end-to-end deep neural network approach to detect translational equivalence between sentences in two different languages. In contrast to previous approaches, which typically rely on multiples models and various word alignment features, by leveraging continuous vector representation of sentences we remove the need of any domain specific feature engineering. Using a siamese bidirectional recurrent neural networks, our results against a strong baseline based on a state-of-the-art parallel sentence extraction system show a significant improvement in both the quality of the extracted parallel sentences and the translation performance of statistical machine translation systems. We believe this study is the first one to investigate deep learning for the parallel sentence extraction task. 
\end{abstract}

\section{Introduction}
\label{sect:introduction}
Parallel corpora are a prerequisite for many multilingual natural language processing applications. As they are an invaluable resource, the limited amount of parallel data, which is only available for a relatively small number of language pairs on very few specific domains, is problematic for scaling natural language processing applications. For example, parallel corpora plays a critical role in machine translation since only the words appearing in the vocabulary of the training set can be translated. Thus, there is a growing interest to collect more parallel data, especially for low-resource languages. With the increasing amount of content-related multilingual articles on the World Wide Web, a potential solution to alleviate the parallel data sparsity issue is to identify and extract parallel sentences from this abundant source of information. Consequently, the objective of parallel sentence extraction is to build parallel corpora by extracting parallel sentence pairs from such multilingual articles. They are widely available on the Web for several language pairs and cover various application domains. Among the different multilingual resource, Wikipedia, an online collaborative encyclopedia, is likely the largest repository of comparable corpora in many languages. Comparable corpora can be defined as collections of topic-aligned but non-sentence-aligned multilingual documents. Several recent works have used Wikipedia as a source of data to create high-quality comparable corpora~\cite{Otero:2010,Patry:2011,Cristina:2015} and various parallel sentence extraction systems have been developed over the years to generate new parallel corpora~\cite{Fung:2004,Munteanu:2005,Adafre:2006,Rauf:2009,Smith:2010,Uszkoreit:2010}.

Recent advances in deep learning architectures with recurrent neural networks (RNN) have shown that they can successfully learn complex mapping from variable-length sequences to continuous vector representations. While numerous natural language processing tasks have successfully applied those models, ranging from handwriting generation~\cite{Graves:2013}, to image caption generation~\cite{Vinyals:2014} and to machine comprehension~\cite{Hermann:2015}, most of the multilingual efforts have been devoted to machine translation~\cite{Sutskever:2014,Cho:2014}, although more research interests have been recently devoted to multilingual semantic textual similarity\footnote{http://alt.qcri.org/semeval2017/task2/}.

Previous approaches have empirically demonstrated that the inclusion of extracted parallel sentence pairs improved the performance of statistical machine translation (SMT) systems, however such methods rely on a significant amount of feature engineering and are difficult to adapt to out-of-domain contexts. In this paper, we propose a deep neural network approach to parallel sentence extraction that takes as input a pair of documents and outputs sentence pairs classified as translations of each other. Compared to previous approaches which require specialized metadata from document structure or to train multiple different models, our model is learned end-to-end and uses only raw sentence pairs. We show empirically that our proposed approach outperforms a competitive baseline based on the works of~\cite{Munteanu:2005} and~\cite{Smith:2010}. To justify the effectiveness of the proposed approach, we add the sentence pairs extracted from Wikipedia articles to a parallel corpus to train SMT systems and show improvements in BLEU scores. Our experiments show that we can achieve promising results by removing the need of any specific feature engineering or external resources. To the best of our knowledge, this is the first time deep learning is applied to extract parallel sentence pairs.

\section{Related work}
\label{sect:related-work}
A variety of approaches have been developed to extract parallel sentences from comparable corpora. In particular,~\cite{Munteanu:2005} presents a system which relies on a multi-step procedure to extract sentence pairs from comparable corpora of newspaper articles. The procedure needs to align pairs of similar documents using publication dates and an information retrieval system. From each such pair, all possible sentence pairs from the Cartesian product of the two documents are passed through a word-overlap and sentence-length ratio filter to obtain a set of candidate sentence pairs. These candidate sentence pairs are sent to a classifier which determines whether two sentences are translations of each other. By using the extracted sentence pairs as additional training data for SMT systems, they demonstrate that this improves the translation performance.~\cite{Smith:2010} extends this approach by exploiting the structure and metadata of interlanguage linked Wikipedia article pairs and introducing several new features, such as distortion features and others that take into account the position of the current and previously aligned sentences. They use their augmented set of features in a conditional random field and obtain state-of-the-art results.~\cite{Rauf:2009} proposes a simpler approach, in which they use an SMT system built from a small parallel corpus. Instead of using a classifier, they translate the source language side of a comparable corpus to find candidate sentences on the target language side. They determine if a translated source sentence and a candidate target sentence are parallel by measuring the word error rate and the translation error rate. Although~\cite{Cristina:2015} focuses on aligning domain-specific parallel documents from Wikipedia, they compute similarities between sentence pairs by cosine and length factor measures popular in cross-language information retrieval. Even if they obtain relatively low precision and recall scores with their extraction method, they observe that extracted domain-specific sentence pairs significantly improved translation quality of STM systems on in-domain data.

\section{Approach}
\label{sect:approach}
\subsection{Negative Sampling}
For training purpose, we use a parallel corpus \(C\) consisting of \(n\) parallel sentence pairs \((\mathbf{s}^{S}_{k}, \mathbf{s}^{T}_{k})\), for \(k \in \{1,\dots,n\}\), where \(S\) and \(T\) denote the source sentences and target sentences sets. These parallel sentence pairs are the positive examples of our training set. Since we want a model that learns differentiable vector representations to distinguish parallel from non-parallel sentences, we need to generate negative examples. Therefore, at the beginning of each training epoch, for every pair of parallel sentences we randomly sample \(m\) negative sentence pairs \((\mathbf{s}^{S}_{k}, \mathbf{s}^{T}_{j})\), for \(j \neq k\).\footnote{In any parallel corpus there might be many redundant and similar sentence pairs. Thus, relying only on randomness to select negative sentence pairs does not guarantee that a sampled sentence pair is truly negative and might occasionally generate false negatives.} Hence, for each epoch our training data consists of \(n(1 + m)\) triples \((\mathbf{s}^{S}_{i}, \mathbf{s}^{T}_{i}, y_{i})\), where \(\mathbf{s}^{S}_{i} = (w^{S}_{i,1}, \dots, w^{S}_{i,N})\) is a source sentence of \(N\) tokens, \(\mathbf{s}^{T}_{i} = (w^{T}_{i,1}, \dots, w^{T}_{i,M})\) is a target sentence of \(M\) tokens, and \(y_{i}\) is the label representing the translation relationship between \(\mathbf{s}^{S}_{i}\) and \(\mathbf{s}^{T}_{i}\), so that \(y_{i} = 1\) if \((\mathbf{s}^{S}_{i}, \mathbf{s}^{T}_{i}) \in C\) and \(y_{i} = 0\) otherwise. 

\subsection{Model}
\begin{figure}[t]
  \includegraphics[width=\linewidth]{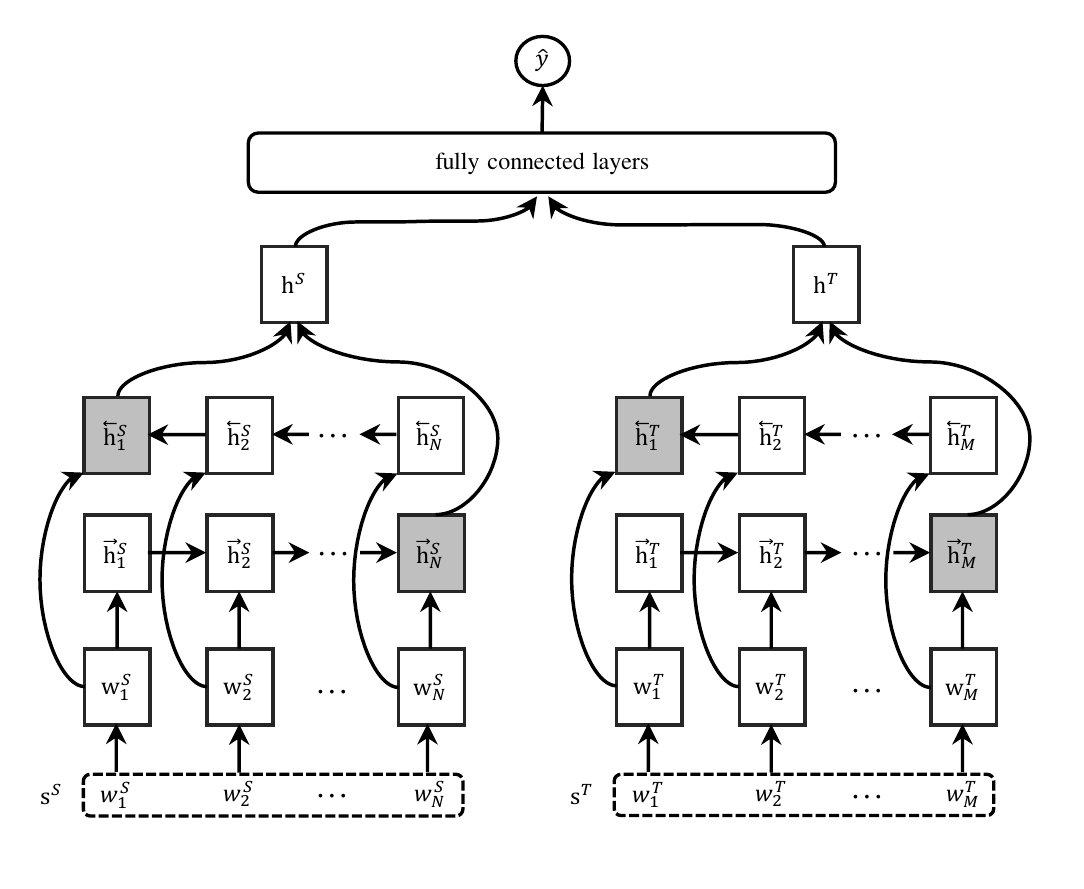}
  \caption{Architecture for the siamese bidirectional recurrent neural networks. The final recurrent state of the forward and backward networks are concatenated and then fed into fully connected layers culminating in a sigmoid layer.}
  \label{fig:model}
\end{figure}

Our idea is to use deep neural networks to learn cross-language semantics between sentence pairs to estimate the probability that they are translations of each other, \(p(y_{i}=1|\mathbf{s}^{S}_{i},\mathbf{s}^{T}_{i})\). The proposed model architecture is a siamese network~\cite{Bromley:93} consisting of a bidirectional RNN (BiRNN)~\cite{Schuster:97}) sentence encoder with recurrent activation functions such as long short-term memory units (LSTM)~\cite{Hochreiter:97} or gated recurrent units (GRU)~\cite{Cho:2014}. Since we want vector representations in a shared vector space we use a siamese network with tied weights. As illustrated in Figure~\ref{fig:model}, our architecture uses a shared BiRNN sentence encoder that outputs a vector representation for the source and target sentences.

To avoid repetition and for clarity, we only define equations of the BiRNN encoding the source sentence. For the target sentence, simply substitute \(S\) for \(T\). At each time step \(t\), the token in the \(i\)-th sentence, \(w^{S}_{i,t}\), defined by its integer index \(k\) in the vocabulary \(V^{S}\), is represented as a one-hot vector \(\mathbf{w}_{k} \in \R^{|V^{S}|}\) whose \(k\)-th element is 1 and all other elements are 0. The one-hot vector is multiplied with a learned embedding matrix \(\mathbf{E}^{S} \in \R^{|V^{S}| \times d_{e}}\) to get a continuous vector representation (word embedding) \(\mathbf{w}^{S}_{i, t} \in \R^{d_{e}}\), which serves as input for the forward and backward recurrent states in the BiRNN encoder, \(\overrightarrow{\mathbf{h}}^{S}_{i, t}\) and \(\overleftarrow{\mathbf{h}}^{S}_{i, t}\). The forward RNN reads the variable-length sentence and updates its recurrent state from the first token until the last one to create a fixed-size continuous vector representation of the sentence, \(\mathbf{h}^{S}_{i, N} \in \R^{d_{h}}\). The backward RNN processes the sentence in reverse. In our experiments, we use the concatenation of the last recurrent state in both directions as a final representation \(\mathbf{h}^{S}_{i}\ = [\overrightarrow{\mathbf{h}}^{S}_{i,N}\ ; \overleftarrow{\mathbf{h}}^{S}_{i,1}]\) (see Figure~\ref{fig:model})\footnote{We considered combining the recurrent states with average pooling and max pooling to obtain a fixed-size vector representation, but obtained inferior performance.}.
\begin{gather}
  \mathbf{w}^{S}_{i,t} = \mathbf{E}^{S^\top} \mathbf{w}_{k} \\
  \overrightarrow{\mathbf{h}}^{S}_{i,t} = \phi(\overrightarrow{\mathbf{h}}^{S}_{i, t-1},
    \mathbf{w}^{S}_{i,t}) \\
  \overleftarrow{\mathbf{h}}^{S}_{i,t} = \phi(\overleftarrow{\mathbf{h}}^{S}_{i, t+1},
    \mathbf{w}^{S}_{i,t})
\end{gather}
\(\phi\) can be any recurrent activation function, such as LSTM or GRU. After both source and target sentences have been encoded, we capture their matching information by using their element-wise product and absolute element-wise difference. We estimate the probability that the sentences are translations of each other by feeding the matching vectors into fully connected layers:
\begin{gather}
  \mathbf{h}_{i}^{(1)} = \mathbf{h}^{S}_{i} \odot \mathbf{h}^{T}_{i} \\
  \mathbf{h}_{i}^{(2)} = |\mathbf{h}^{S}_{i} - \mathbf{h}^{T}_{i}| \\
  \mathbf{h}_{i} = tanh(\mathbf{W}^{(1)}\mathbf{h}_{i}^{(1)} + \mathbf{W}^{(2)}\mathbf{h}_{i}^{(2)} + \mathbf{b}) \\
  p(y_{i}=1|\mathbf{h}_{i}) = \sigma(\mathbf{W}^{(3)}\mathbf{h}_{i} + c)
\end{gather}
where \(\sigma\) is the sigmoid function, \(\mathbf{W}^{(1)}\), \(\mathbf{W}^{(2)}\), \(\mathbf{W}^{(3)}\), \(\mathbf{b}\) and \(c\) are model parameters. The model is trained by minimizing the cross entropy of our labeled sentence pairs:
\begin{equation}
  \begin{split}
   \mathcal{L}
   = &-\sum^{n(1+m)}_{i=1} y_{i} \log \sigma(\mathbf{W}^{(3)}\mathbf{h}_{i} + c) \\
     &-(1-y_{i}) \log (1-\sigma(\mathbf{W}^{(3)}\mathbf{h}_{i} + c))
  \end{split}
\end{equation}

For prediction, a sentence pair is classified as parallel if the probability score is greater than or equal to a decision threshold \(\rho\) that we need to fix.
\begin{equation}
  \hat{y}_{i} =
    \begin{cases}
      1 & \text{if}\ p(y_{i}=1|\mathbf{h}_{i}) \geq \rho \\
      0 & \text{otherwise}
    \end{cases}
\end{equation}

\section{Experiments}
\label{sect:experiments}
To assess the effectiveness of our approach we compare it in different settings against the baseline model described in Section~\ref{ssec:baseline}. First, we measure the precision, recall and F\(_{1}\) scores by extracting parallel sentences from a standard parallel corpus in Section~\ref{ssec:model-evaluation}. To compare the approaches with pseudo comparable corpora with different degrees of comparability, we insert noisy non-parallel sentences into the parallel corpus. In Section~\ref{ssec:machine-translation}, we extract sentence pairs from real comparable corpora and validate their utility by measuring their impact on SMT systems.

\subsection{Evaluation metrics}
\label{ssec:metrics}
For the evaluation of the performance of our models, a sentence pair predicted as parallel is correct if it is present in the parallel sentence pairs of the dataset. Precision is the proportion of truly parallel sentence pairs among all extracted sentence pairs. Recall is the proportion of truly parallel extracted sentence pairs among all parallel sentence pairs in the dataset. The F\(_{1}\) score is the harmonic mean of precision and recall.

For the statistical machine translation evaluation we use the BLEU score~\cite{Papineni:2002} as an evaluation metric using the multi-bleu script from Moses~\cite{Moses:2007}\footnote{https://github.com/moses-smt/mosesdecoder}.

\subsection{Datasets}
\label{ssec:datasets}
The most reliable way to compare the precision, recall and F\(_{1}\) scores would be to have professional translators manually annotate parallel sentences from comparable corpora. However, this option is expensive and impractical. Therefore, for this task it is common practice to compare different approaches using aligned texts from known parallel corpora. Thus, to compute our evaluation metrics we use the WMT'15 English to French datasets\footnote{http://www.statmt.org/wmt15/translation-task.html}. Our training set consists of 500k parallel sentence pairs randomly selected from the Europarl v7 corpus~\cite{Koehn:2005}. The vocabulary size is 69k for English and 84k for French. We argue that parallel sentence extraction in practice requires domain adaptation, i.e. data during prediction will most probably cover other domains than the ones found in the training set, so we focus on out-of-domain test sets. Therefore, we use the first 1,000 parallel sentence pairs of newstest2012 for the model evaluation experiment. For the STM evaluation experiment, the comparable corpora we use to extract parallel sentences are English-French Wikipedia article pairs from the Wikipedia dumps\footnote{https://dumps.wikimedia.org/} and the test set is newstest2013. Data processing is performed to clean and segment the Wikipedia XML documents into sentences. We normalize and tokenize all datasets with the scripts from Moses. The maximum sentence length is set to 80 tokens.

\begin{figure*}
  \includegraphics[width=\linewidth]{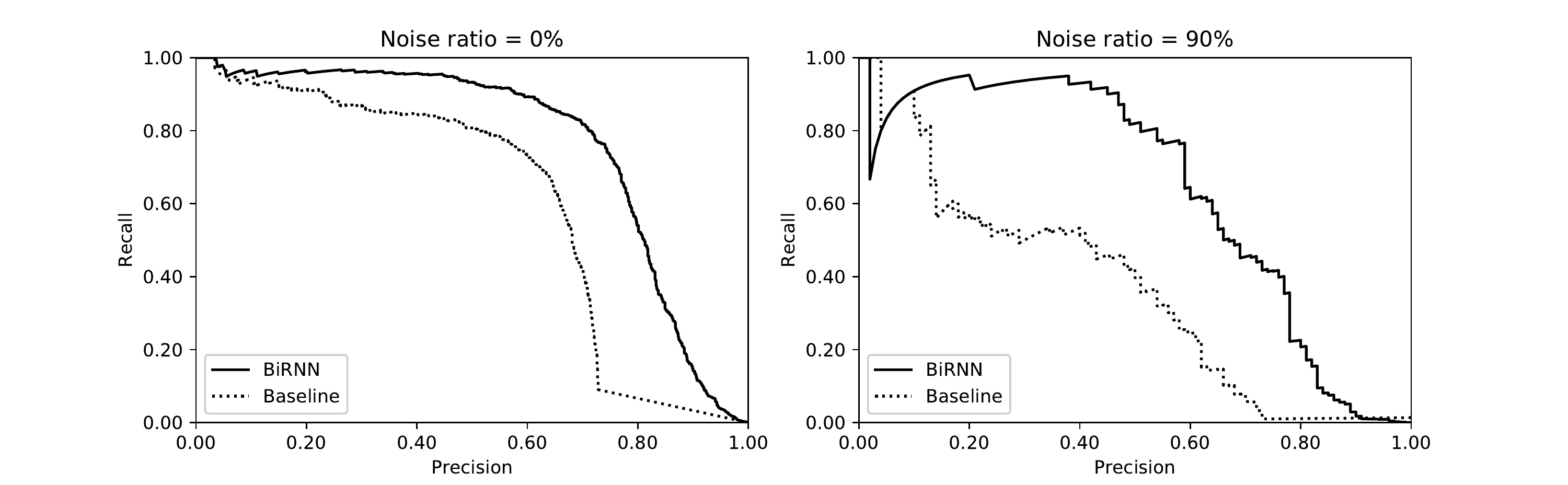}
  \caption{Precision-Recall curve of the models evaluated on the Cartesian product of the 1,000 first sentence pairs of newstest2012 without noise (left) and with a noise ratio of 90\% (right).}
  \label{fig:precision-recall}
\end{figure*}

\subsection{Baseline}
\label{ssec:baseline}
For comparison, we use a parallel sentence extraction system developed in-house by~\cite{Alex:2014} based on the work of~\cite{Munteanu:2005} and~\cite{Smith:2010}. The system consists of a candidate sentence pair filtering process and three models; two word alignment models and a maximum entropy classifier. The word alignment models are trained on both language directions using our training set of 500k parallel sentence pairs. For the classifier, we select another 10k parallel sentence pairs from the held-out Europarl dataset and choose a negative to positive ratio (non-parallel to parallel sentence pairs) to select the number of negative sentence pairs\footnote{Munteanu \& Marcu~\shortcite{Munteanu:2005} use 5k parallel sentence pairs with a negative to positive ratio not greater than 5. By using 10k parallel sentence pairs over 5k we obtained small performance gains, but we did not observe any significant gain by using more than 10k parallel sentence pairs.}. For example, with a negative to positive ratio of 5 we select 50k negative sentence pairs, so for each epoch the classifier is trained on 60k examples.

\noindent\textbf{Word alignment models}\hspace{1mm} The translation and alignment tables are estimated using the HMM alignment model of~\cite{Vogel:1996}. These probability tables are required to measure the value of the many alignment features used in the classifier. To perform word alignment we use an IBM model 2. The translations with a probability score above 10\% from the estimated translation tables are used to infer bilingual dictionaries that are used in the word-overlap filter for candidate sentence pair selection. For our experiments, we use the GIZA++ implementation~\cite{Och:2003}\footnote{http://www.statmt.org/moses/giza/GIZA++.html} to train our word alignment models.

\noindent\textbf{Maximum entropy classifier}\hspace{1mm} The classifier uses 31 features which are based on the work of~\cite{Munteanu:2005} and~\cite{Smith:2010}. They rely on word-level alignment features between two sentences, such as the number and percentage of connected (unconnected) words, the top three largest fertilities, percentage of source words with fertility 1, 2, 3 or more, length of the longest connected (unconnected) substring, log probability of the alignment, and also general features, such as the lengths of the sentences, length difference, length ratio and the percentage of words on each side that have a translation on the other side. A sentence pair is classified as parallel if the classifier outputs a probability score greater than or equal to a decision threshold \(\rho\) which needs to be fixed.

\noindent\textbf{Candidate sentence pair selection}\hspace{1mm} During training, a sentence pair filtering process is used to select a fixed number of negative sentence pairs to train the maximum entropy classifier. It is also used during prediction to filter out the unlikely sentence pairs of the Cartesian product. First, it verifies that the ratio of the lengths of the two sentences is not greater than two. It then uses a word-overlap filter to check for both sentences that at least 50\% of their words have a translation in the other sentence, according to the bilingual dictionaries inferred from the word alignment models. Every pair that do not fulfill these two conditions are discarded.

During our experiments, we observed that the filtering process eliminates more than 99\% of the candidate sentence pairs from the Cartesian product of the test set and that the classifier alone is not able to classify truly parallel sentences. The sentence pair filtering process is only applied to the baseline model.

\subsection{Training settings}
\label{ssec:training}
Our neural network models are implemented using TensorFlow~\cite{Tensorflow:2016}. We use a siamese BiRNN with a single layer in each direction with 512-dimensional word embeddings and 512-dimensional recurrent states. We use GRU as recurrent activation functions since they consistently outperformed LSTM by a small margin in our experiments. The hidden layer of the fully connected layers has 256 hidden units. We initialize all parameters uniformly using TensorFlow's default uniform unit scaling initialization, except for all biases being initialized to zero. To train our models, we use Adam optimizer~\cite{Kingma:2014} with a learning rate of 0.0002 and a minibatch of 128 examples. Models are trained for a total of 15 epochs. To avoid exploding gradients, we apply gradient clipping such that the norm of all gradients is no larger than 5~\cite{Pascanu:2013}. We apply dropout to prevent overfitting with a probability of 0.2 and 0.3 for the non-recurrent input and output connections respectively. Training is performed on a single GPU.

\section{Results}
\label{sect:results}
\subsection{Model Evaluation}
\label{ssec:model-evaluation}
In this section we measure the precision, recall and F\(_{1}\) scores to compare both methods. We use a siamese BiRNN model trained with 7 negative samples and a baseline model trained on a balanced training set with a negative to positive ratio of 1\footnote{We experimented with different number of negative samples and after a value of 7 we observed that the marginal benefit of adding more negative samples is not significant. As for the baseline model, we observed better performance and more stability when it is trained on a balanced training set.}. In order to compare the feasibility of our approach on different degrees of comparability of comparable corpora, we insert noisy non-parallel sentences into the test set by substituting a defined number of target sentences with external target sentences from the held-out sentence pairs of the newstest2012 corpus. For example, with a noise ratio of 60\%, 600 out of the 1,000 sentence pairs are not parallel, such that only 0.04\% of the sentence pairs in the Cartesian product are truly parallel. Figure~\ref{fig:precision-recall} shows the precision-recall curve of the models evaluated on the test set with a noise ratio of 0\% and 90\%. In Table~\ref{table:optimal-f1}, we report the scores at the decision threshold value fixed at the optimal F\(_{1}\) value. We see that we are able to consistently outperform by a significant margin the results obtained with the baseline model. Our approach has a F\(_{1}\) improvement over the baseline model of 9.61\% and 19.61\% on the test set with a noise ratio of 0\% and 90\%, respectively.

\begin{table}[t]
  \fontsize{8}{10}\selectfont
  \centering
  \setlength{\tabcolsep}{1.5pt}
  \begin{tabular}{*{1}{l}*{8}{c}}
    \toprule 
    & \multicolumn{4}{c}{\thead{Noise ratio = 0\%}} & \multicolumn{4}{c}{\thead{Noise ratio = 90\%}} \\
    \cmidrule(lr){2-5} \cmidrule(lr){6-9}
    \thead{Model} 
    & \multicolumn{1}{c}{\thead{P (\%)}} & \multicolumn{1}{c}{\thead{R (\%)}}
    & \multicolumn{1}{c}{\thead{F\(_{1}\) (\%)}} & \multicolumn{1}{c}{\thead{\(\boldsymbol{\rho}\)}}
    & \multicolumn{1}{c}{\thead{P (\%)}} & \multicolumn{1}{c}{\thead{R (\%)}}
    & \multicolumn{1}{c}{\thead{F\(_{1}\) (\%)}} & \multicolumn{1}{c}{\thead{\(\boldsymbol{\rho}\)}} \\
    \midrule
    \addlinespace[5pt]
    BiRNN & \textbf{83.0} & \textbf{69.6} & \textbf{75.7} & 0.99 & \textbf{70.6} & \textbf{59.0} & \textbf{66.7} & 0.99 \\
    Baseline & 73.1 & 60.3 & 66.1 & 0.85 & 46.2 & 48.0 & 47.1 & 0.97 \\ 
    \bottomrule
  \end{tabular}
  \caption{Precision, recall and F\(_{1}\) scores at decision threshold value \(\rho\) maximizing the F\(_{1}\) score on our test set.}
  \label{table:optimal-f1}
\end{table}

\begin{figure}[t]
  \includegraphics[width=\linewidth]{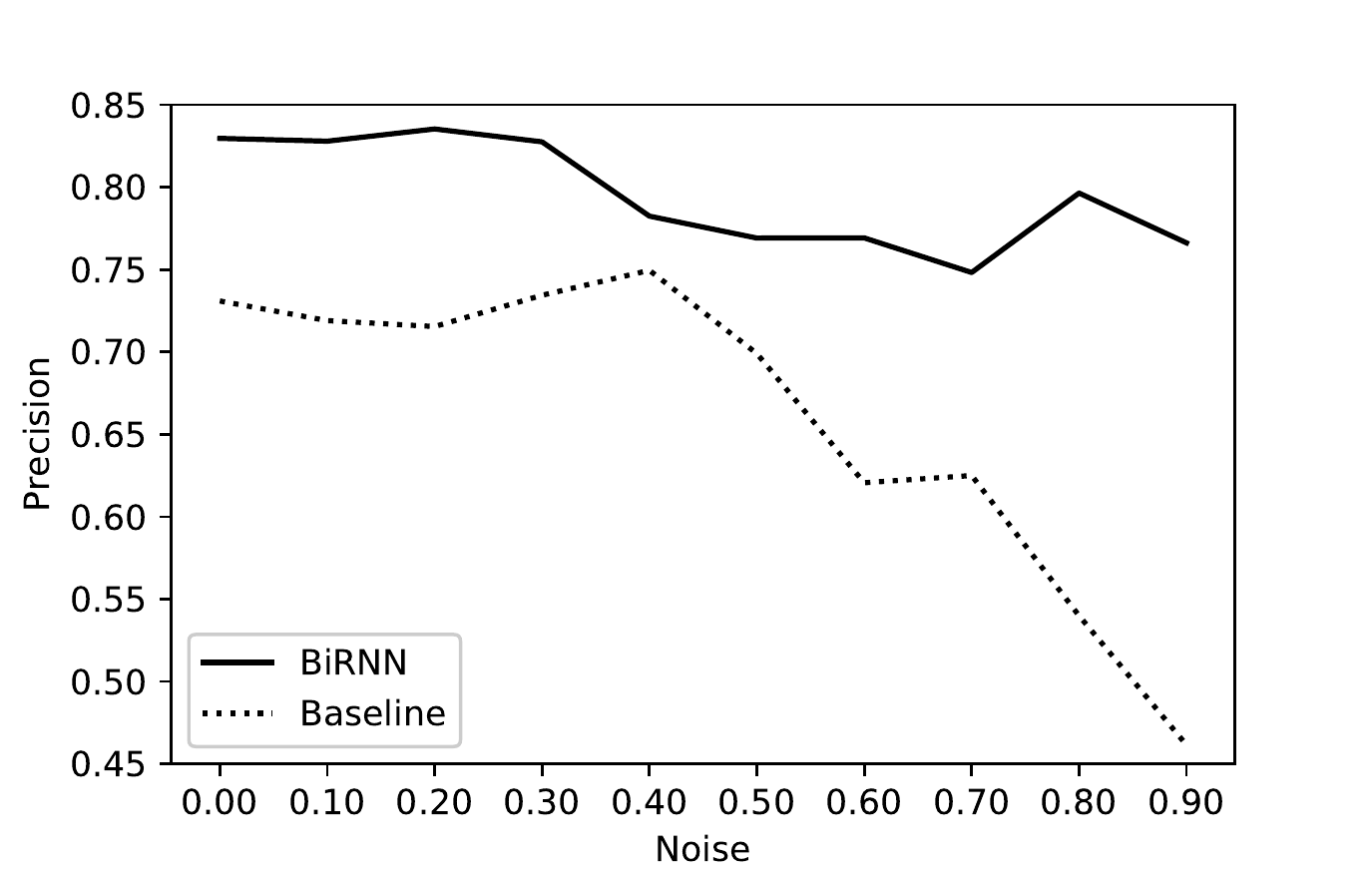}
  \caption{Precision score as the number of noisy non-parallel sentences in the test set increases.}
  \label{fig:precision-noise}
\end{figure}

\begin{figure}[t]
  \includegraphics[width=\linewidth]{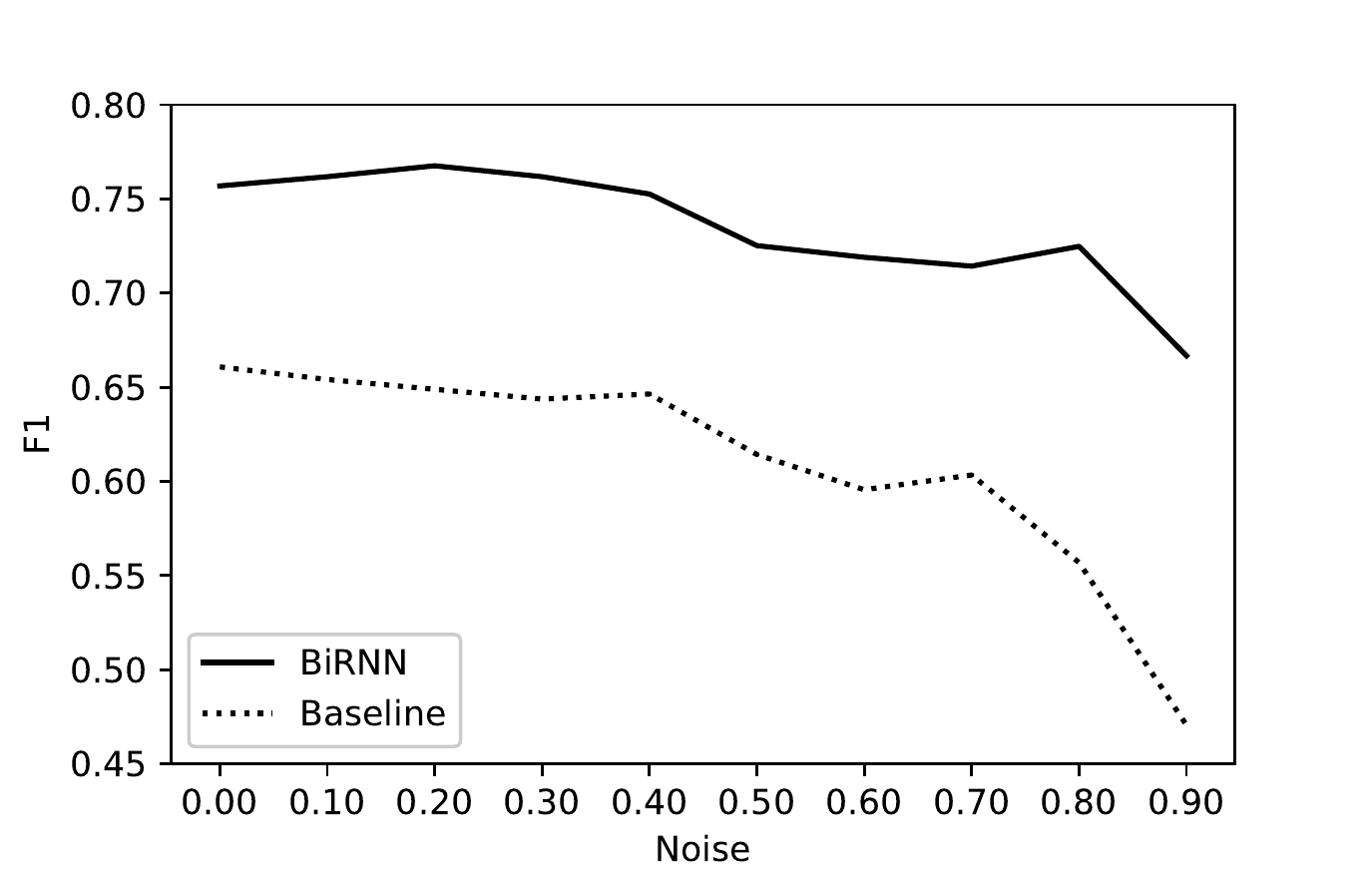}
  \caption{F\(_{1}\) score as the number of noisy non-parallel sentences in the test set increases.}
  \label{fig:f1-noise}
\end{figure}

\begin{figure}[!ht]
  \includegraphics[width=\linewidth]{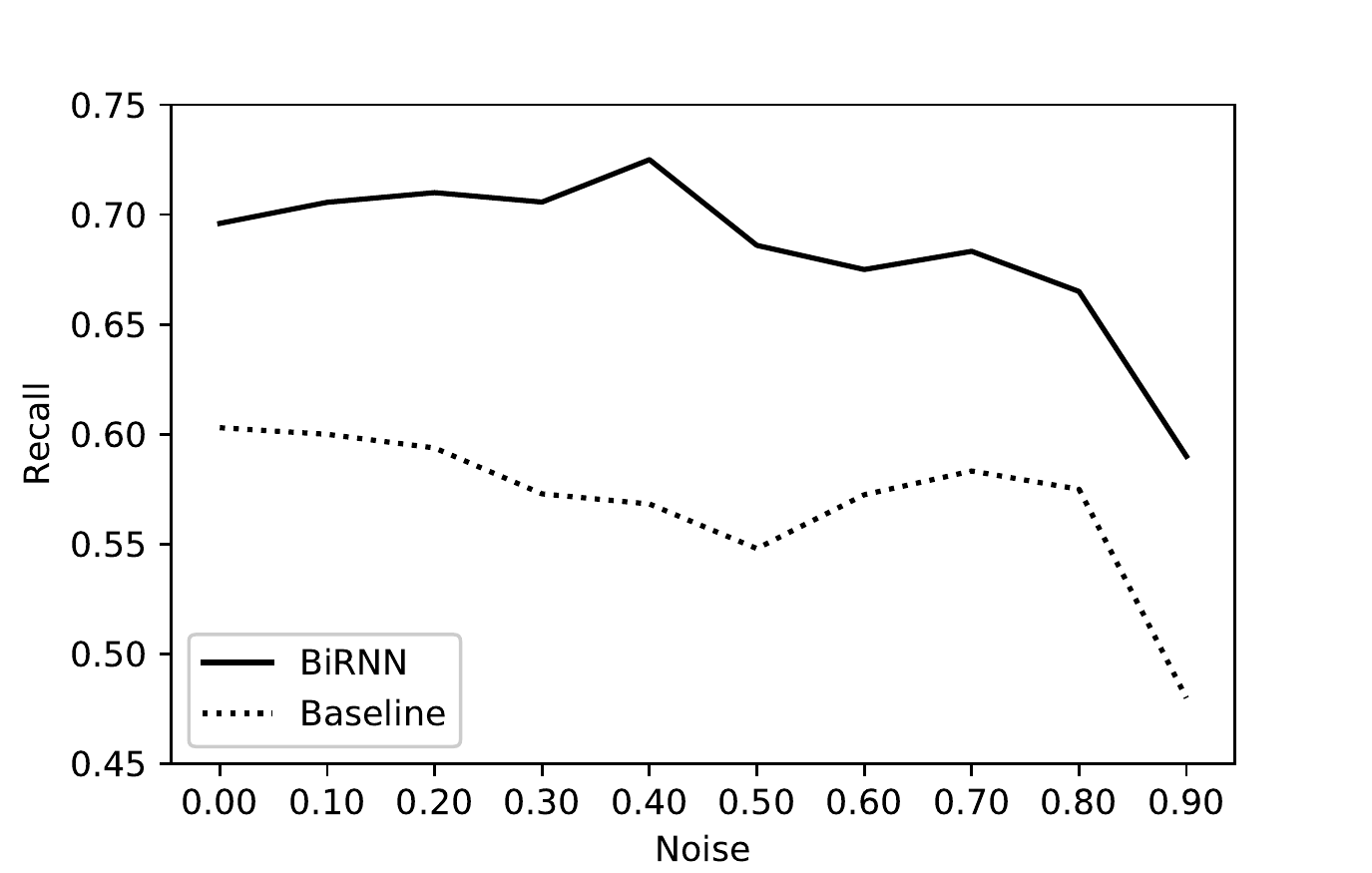}
  \caption{Recall score as the number of noisy non-parallel sentences in the test set increases.}
  \label{fig:recall-noise}
\end{figure}

In Figures~\ref{fig:precision-noise},~\ref{fig:recall-noise} and~\ref{fig:f1-noise}, we compare the precision, recall and F\(_{1}\) scores as the noise ratio in our test set increases. We observe that it becomes harder to identify parallel sentences as the number of non-parallel sentences increases in the test set. However, we see that our neural network-based approach outperforms the baseline model across the line. In contrast to the baseline, the performance of our method stays relatively stable and starts to degrade at very high noise ratios. Within that range, we believe it is more representative to document pairs found in real comparable corpora. While we present the scores for the sentence pairs extracted with a decision threshold value fixed at the optimal F\(_{1}\) score, some may believe that the precision of the extracted pairs is more important than the recall and that having an approach with the best F\(_{1}\) score is not optimal. In this regard,~\cite{Goutte:2012} finds that SMT systems are robust to noise in training data and that recall can be even more important than precision\footnote{However, it might not be the case for neural machine translation systems based on distributional semantic representations where precision could be the score to prioritize. We need to further investigate the impact and leave it for future work.}. In any case, we see that our approach gives a better precision at a larger recall value, meaning that setting the decision threshold in order to obtain a desired precision 
\noindent will lead to a larger number of high-quality parallel sentences. The value of the decision threshold has a direct impact on the quality and the amount of extracted sentence pairs. In our case, because we are in presence of datasets with highly imbalanced classes, we recommend to use a very high value \(\rho \approx 0.99\) to reduce the number of false positives.

\subsection{Statistical Machine Translation Evaluation}
\label{ssec:machine-translation}

\begin{table}[t]
  \centering
  \fontsize{9}{11}\selectfont
  \setlength{\tabcolsep}{4pt}
  \begin{tabular}{*{2}{l}*{1}{r}*{2}{l}}
    \toprule 
    \thead{Language} & \thead{Model} & \thead{Sentences} & \thead{Tokens} & \thead{Length} \\
    \midrule
    \multirow{2}{*}{English} & BiRNN & 1,487,769 & 29,740,242 & 20 \(\pm\) 11 \\
    					     & Baseline & 792,514 & 14,310,191 & 18 \(\pm\) 9 \\
    \midrule
    \multirow{2}{*}{French} & BiRNN  & 1,487,769 & 32,613,325 & 22 \(\pm\) 12 \\ 
    			      	    & Baseline & 792,514 & 15,245,228 & 19 \(\pm\) 10 \\
    \bottomrule
  \end{tabular}
  \caption{Statistics of the size of the parallel corpora extracted from the English-French Wikipedia article pairs. Length is the average and standard deviation of the number of tokens in the sentences.}
  \label{table:corpus-size}
\end{table}

The objective of parallel sentence extraction is to increase the size of existing parallel corpora and to broaden the covered domains in order to improve the generalization of machine translation systems. To justify the utility of our proposed approach, we extract parallel sentences from interlanguage linked English-French Wikipedia articles and evaluate their quality by measuring the BLEU scores on SMT systems. We want a good balance between the quality and the number of extracted sentence pairs, so for each approach we set the decision threshold value \(\rho\) equal to the value maximizing the F\(_{1}\) score with a noise ratio of 90\% (see Table~\ref{table:optimal-f1}). We use these values as a rough estimate to represent the degree of comparability present in Wikipedia article pairs.

For both methods, the classifier independently classifies each \(i\)-th sentence pair as parallel if \(\hat{y}_{i} \geq \rho\). This can lead to a situation where a source sentence is paired to several target sentences, or vice versa. To guarantee that sentences in both languages appear at most in a single pair (i.e. one-to-one alignment), as a post-treatment step we employ a greedy strategy that sorts the extracted sentence pairs by best probability score and greedily iterates over this sequence by eliminating pairs whose source or target sentence has been already paired. Information on the size of the extracted parallel corpora, keeping only the sentence pairs in which both sentences contain at least 3 tokens, is presented in Table~\ref{table:corpus-size}. As we expected, with the superior precision and recall values of our approach (see Section~\ref{ssec:model-evaluation}), the BiRNN extracts more sentence pairs than the baseline. In fact, there is a quality-size trade-off and it is possible to set the \(\rho\) value in order maximize the quality (size) of the extracted parallel corpus, to the detriment of its size (quality). We calculated the coverage ratio and we found that 77\% of the sentence pairs extracted by the baseline model were also extracted from our BiRNN approach.

To train the phrase-based translation systems~\cite{Koehn:2003} we use the Moses toolkit. As baseline SMT system, we train an SMT system on our training set consisting of 500k parallel sentence pairs selected from the Europarl corpus described in Section~\ref{ssec:datasets}. We train two additional SMT systems by augmenting the training set with the extracted sentence pairs from the BiRNN and baseline models. Each system uses newstest2012 as tuning set and is evaluated on the English-French translation quality on newstest2013. Since both extracted parallel corpus are not on the same order of magnitude, we sorted the sentence pairs by similarity score in descending order and used the top 500k sentence pairs to train new SMT systems with training sets of equal size. Table~\ref{table:bleu-scores} shows the BLEU scores for the different SMT systems. When using the full extracted parallel corpus, we see that our approach improves the BLEU score over the baseline SMT system trained solely on the 500k sentence pairs of the Europarl corpus by 4.7 and about 0.8 for the system trained with the extra sentence pairs extracted with the baseline model. When we only use the top 500k sentence pairs with higher similarity score, our approach is on par with the baseline system. Thus, it confirms the quality of the top 500k ranked sentence pairs, such that we could lower the decision threshold value to extract more high-quality parallel sentences. Given the out-of-domain nature of our Wikipedia articles with respect to our training set, those results are encouraging because they show that our approach should adapt well on comparable corpora with a lower degree of comparability. 

\begin{table}[t]
  \centering
  \fontsize{9}{11}\selectfont
  \setlength{\tabcolsep}{5.5pt}
  \begin{tabular}{*{2}{l}*{1}{c}*{1}{r}}
    \toprule 
    \thead{Training Data} & \thead{Model} & \thead{BLEU} & \thead{Sentences} \\
    \midrule
    Europarl & & 21.5 & 500,000 \\ 
    \midrule
    \multirow{2}{*}{+Full}  & BiRNN & 26.2 (+4.7) & 1,987,769 \\
    					   & Baseline & 25.4 (+3.9) & 1,292,514 \\
    \midrule
    \multirow{2}{*}{+Top500k} & BiRNN & 25.0 (+3.5) & 1,000,000 \\ 
    			      	      & Baseline & 24.9 (+3.4) & 1,000,000 \\
    \bottomrule
  \end{tabular}
  \caption{BLEU scores obtained on the newstest2013 test set. Sentences is the number of sentences used to train the SMT systems. The Europarl row is the baseline SMT system trained on 500k sentences pairs from the Europarl corpus.}
  \label{table:bleu-scores}
\end{table}

\section{Discussion}
\label{sect:discussion}
In this work, we presented a deep neural network approach to extract parallel sentences. Our work showed that our approach outperforms by a significant margin a strong baseline based on state-of-the-art parallel sentence extraction system. Traditional systems need to train multiples models and to apply a two-step classification procedure. In contrast, we propose a simpler approach that only requires a parallel corpus to encode sentence pairs in a siamese BiRNN encoder using LSTM or GRU activation functions.

Our work enables exploration for researchers who want to apply more advanced deep learning architectures to the parallel sentence extraction task. We believe our approach is scalable and flexible with different languages or domains. That being said, it would be natural to extend the approach using multiple language pairs. Currently, we do not handle the unknown out-of-vocabulary words, which might be an issue. Although we have evaluated our approach in an out-of-domain setting (training done on the Europarl corpus, extracting parallel sentences from Wikipedia articles and testing on newstest2013) with promising results, we need to further investigate the impact it might have.

We saw that the degree of comparability in a pair of documents negatively impact the performance of our approach. A more advanced analysis on the hyperparameters settings could be applied to improve the generalization of the model. Instead of only selecting random negative samples, a promising next step could be to use a mix of random and hard negative samples in our training set (i.e. similar non-parallel sentence pairs). However, to achieve that we are forced to do an extra feedforward pass over the whole training set at the beginning of each training epoch to obtain the similar non-parallel sentences, otherwise we need external resources, such as bilingual dictionaries or pre-trained word embeddings. The data and our code are available on github.

\bibliography{acl2017}

\begin{thebibliography}{}
\expandafter\ifx\csname natexlab\endcsname\relax\def\natexlab#1{#1}\fi

\bibitem[{Abadi et~al.(2016)Abadi, Barham, Chen, Chen, Davis, Dean, Devin,
  Ghemawat, Irving, Isard, Kudlur, Levenberg, Monga, Moore, Murray, Steiner,
  Tucker, Vasudevan, Warden, Wicke, Yu, and Zheng}]{Tensorflow:2016}
Martin Abadi, Paul Barham, Jianmin Chen, Zhifeng Chen, Andy Davis, Jeffrey
  Dean, Matthieu Devin, Sanjay Ghemawat, Geoffrey Irving, Michael Isard,
  Manjunath Kudlur, Josh Levenberg, Rajat Monga, Sherry Moore, Derek~G. Murray,
  Benoit Steiner, Paul Tucker, Vijay Vasudevan, Pete Warden, Martin Wicke, Yuan
  Yu, and Xiaoqiang Zheng. 2016.
\newblock Tensorflow: A system for large-scale machine learning.
\newblock In {\em 12th USENIX Symposium on Operating Systems Design and
  Implementation (OSDI 16)\/}. pages 265--283.

\bibitem[{Abdul-Rauf and Schwenk(2009)}]{Rauf:2009}
Sadaf Abdul-Rauf and Holger Schwenk. 2009.
\newblock \href{http://dl.acm.org/citation.cfm?id=1609067.1609068}{On the use
  of comparable corpora to improve smt performance}.
\newblock In {\em Proceedings of the 12th Conference of the European Chapter of
  the Association for Computational Linguistics\/}. Association for
  Computational Linguistics, Stroudsburg, PA, USA, EACL '09, pages 16--23.
\newblock
  \href{http://dl.acm.org/citation.cfm?id=1609067.1609068}{http://dl.acm.org/citation.cfm?id=1609067.1609068}.

\bibitem[{Adafre and de~Rijke(2006)}]{Adafre:2006}
Sisay Adafre and Maarten de~Rijke. 2006.
\newblock Finding similar sentences across multiple languages in wikipedia.

\bibitem[{Barr{\'o}n-Cede{\~{n}}o et~al.(2015)Barr{\'o}n-Cede{\~{n}}o,
  Espa{\~{n}}a-Bonet, Boldoba, and M{\`a}rquez}]{Cristina:2015}
Alberto Barr{\'o}n-Cede{\~{n}}o, Cristina Espa{\~{n}}a-Bonet, Josu Boldoba, and
  Llu{\'i}s M{\`a}rquez. 2015.
\newblock \href{https://doi.org/10.18653/v1/W15-3402}{A factory of comparable
  corpora from wikipedia}.
\newblock Association for Computational Linguistics, pages 3--13.
\newblock
  \href{https://doi.org/10.18653/v1/W15-3402}{https://doi.org/10.18653/v1/W15-3402}.

\bibitem[{B\'{e}rard(2014)}]{Alex:2014}
Alexandre B\'{e}rard. 2014.
\newblock Better handling of a bilingual collection of texts.
\newblock MSc thesis, Universit\'{e} de Montr\'{e}al.

\bibitem[{Bromley et~al.(1994)Bromley, Guyon, LeCun, S\"{a}ckinger, and
  Shah}]{Bromley:93}
Jane Bromley, Isabelle Guyon, Yann LeCun, Eduard S\"{a}ckinger, and Roopak
  Shah. 1994.
\newblock
  \href{http://papers.nips.cc/paper/769-signature-verification-using-a-siamese-time-delay-neural-network.pdf}{Signature
  verification using a "siamese" time delay neural network} pages 737--744.
\newblock
  \href{http://papers.nips.cc/paper/769-signature-verification-using-a-siamese-time-delay-neural-network.pdf}{http://papers.nips.cc/paper/769-signature-verification-using-a-siamese-time-delay-neural-network.pdf}.

\bibitem[{Cho et~al.(2014)Cho, van Merrienboer, G{\"{u}}l{\c{c}}ehre, Bougares,
  Schwenk, and Bengio}]{Cho:2014}
Kyunghyun Cho, Bart van Merrienboer, {\c{C}}aglar G{\"{u}}l{\c{c}}ehre, Fethi
  Bougares, Holger Schwenk, and Yoshua Bengio. 2014.
\newblock \href{http://arxiv.org/abs/1406.1078}{Learning phrase representations
  using rnn encoder-decoder for statistical machine translation}.
\newblock {\em CoRR\/} abs/1406.1078.
\newblock
  \href{http://arxiv.org/abs/1406.1078}{http://arxiv.org/abs/1406.1078}.

\bibitem[{Fung and Cheung(2004)}]{Fung:2004}
Pascale Fung and Percy Cheung. 2004.
\newblock Mining very non-parallel corpora: Parallel sentence and lexicon
  extraction via bootstrapping and em.
\newblock In {\em Proceedings of EMNLP\/}. pages 57--63.

\bibitem[{Goutte et~al.(2012)Goutte, Carpuat, and Foster}]{Goutte:2012}
Cyril Goutte, Marine Carpuat, and Georges Foster. 2012.
\newblock
  \href{http://www-labs.iro.umontreal.ca/~foster/papers/align-error-amta12.pdf}{The
  impact of sentence alignment errors on phrase-based machine translation
  performance}.
\newblock In {\em The Association for Machine Translation in the Americas
  2012\/}. AMTA '12.
\newblock
  \href{http://www-labs.iro.umontreal.ca/~foster/papers/align-error-amta12.pdf}{http://www-labs.iro.umontreal.ca/~foster/papers/align-error-amta12.pdf}.

\bibitem[{Graves(2013)}]{Graves:2013}
Alex Graves. 2013.
\newblock \href{http://arxiv.org/abs/1308.0850}{Generating sequences with
  recurrent neural networks}.
\newblock {\em CoRR\/} abs/1308.0850.
\newblock
  \href{http://arxiv.org/abs/1308.0850}{http://arxiv.org/abs/1308.0850}.

\bibitem[{Hermann et~al.(2015)Hermann, Kocisky, Grefenstette, Espeholt, Kay,
  Suleyman, and Blunsom}]{Hermann:2015}
Karl~Moritz Hermann, Tomas Kocisky, Edward Grefenstette, Lasse Espeholt, Will
  Kay, Mustafa Suleyman, and Phil Blunsom. 2015.
\newblock
  \href{http://papers.nips.cc/paper/5945-teaching-machines-to-read-and-comprehend.pdf}{Teaching
  machines to read and comprehend}.
\newblock In C.~Cortes, N.~D. Lawrence, D.~D. Lee, M.~Sugiyama, and R.~Garnett,
  editors, {\em Advances in Neural Information Processing Systems 28\/}, Curran
  Associates, Inc., pages 1693--1701.
\newblock
  \href{http://papers.nips.cc/paper/5945-teaching-machines-to-read-and-comprehend.pdf}{http://papers.nips.cc/paper/5945-teaching-machines-to-read-and-comprehend.pdf}.

\bibitem[{Hochreiter and Schmidhuber(1997)}]{Hochreiter:97}
Sepp Hochreiter and J\"{u}rgen Schmidhuber. 1997.
\newblock \href{https://doi.org/10.1162/neco.1997.9.8.1735}{Long short-term
  memory}.
\newblock {\em Neural Comput.\/} 9(8):1735--1780.
\newblock
  \href{https://doi.org/10.1162/neco.1997.9.8.1735}{https://doi.org/10.1162/neco.1997.9.8.1735}.

\bibitem[{Kingma and Ba(2014)}]{Kingma:2014}
Diederik~P. Kingma and Jimmy Ba. 2014.
\newblock \href{http://arxiv.org/abs/1412.6980}{Adam: {A} method for stochastic
  optimization}.
\newblock {\em CoRR\/} abs/1412.6980.
\newblock
  \href{http://arxiv.org/abs/1412.6980}{http://arxiv.org/abs/1412.6980}.

\bibitem[{Koehn(2005)}]{Koehn:2005}
Philipp Koehn. 2005.
\newblock \href{http://mt-archive.info/MTS-2005-Koehn.pdf}{{Europarl: A
  Parallel Corpus for Statistical Machine Translation}}.
\newblock In {\em {Conference Proceedings: the tenth Machine Translation
  Summit}\/}. AAMT, AAMT, Phuket, Thailand, pages 79--86.
\newblock
  \href{http://mt-archive.info/MTS-2005-Koehn.pdf}{http://mt-archive.info/MTS-2005-Koehn.pdf}.

\bibitem[{Koehn et~al.(2007)Koehn, Hoang, Birch, Callison-Burch, Federico,
  Bertoldi, Cowan, Shen, Moran, Zens, Dyer, Bojar, Constantin, and
  Herbst}]{Moses:2007}
Philipp Koehn, Hieu Hoang, Alexandra Birch, Chris Callison-Burch, Marcello
  Federico, Nicola Bertoldi, Brooke Cowan, Wade Shen, Christine Moran, Richard
  Zens, Chris Dyer, Ond\v{r}ej Bojar, Alexandra Constantin, and Evan Herbst.
  2007.
\newblock \href{http://dl.acm.org/citation.cfm?id=1557769.1557821}{Moses: Open
  source toolkit for statistical machine translation}.
\newblock In {\em Proceedings of the 45th Annual Meeting of the ACL on
  Interactive Poster and Demonstration Sessions\/}. Association for
  Computational Linguistics, Stroudsburg, PA, USA, ACL '07, pages 177--180.
\newblock
  \href{http://dl.acm.org/citation.cfm?id=1557769.1557821}{http://dl.acm.org/citation.cfm?id=1557769.1557821}.

\bibitem[{Koehn et~al.(2003)Koehn, Och, and Marcu}]{Koehn:2003}
Philipp Koehn, Franz~Josef Och, and Daniel Marcu. 2003.
\newblock \href{https://doi.org/10.3115/1073445.1073462}{Statistical
  phrase-based translation}.
\newblock In {\em Proceedings of the 2003 Conference of the North American
  Chapter of the Association for Computational Linguistics on Human Language
  Technology - Volume 1\/}. Association for Computational Linguistics,
  Stroudsburg, PA, USA, NAACL '03, pages 48--54.
\newblock
  \href{https://doi.org/10.3115/1073445.1073462}{https://doi.org/10.3115/1073445.1073462}.

\bibitem[{Munteanu and Marcu(2005)}]{Munteanu:2005}
Dragos~Stefan Munteanu and Daniel Marcu. 2005.
\newblock \href{https://doi.org/10.1162/089120105775299168}{Improving machine
  translation performance by exploiting non-parallel corpora}.
\newblock {\em Computational Linguistics\/} 31(4):477--504.
\newblock
  \href{https://doi.org/10.1162/089120105775299168}{https://doi.org/10.1162/089120105775299168}.

\bibitem[{Och and Ney(2003)}]{Och:2003}
Franz~Josef Och and Hermann Ney. 2003.
\newblock \href{https://doi.org/10.1162/089120103321337421}{A systematic
  comparison of various statistical alignment models}.
\newblock {\em Comput. Linguist.\/} 29(1):19--51.
\newblock
  \href{https://doi.org/10.1162/089120103321337421}{https://doi.org/10.1162/089120103321337421}.

\bibitem[{Otero and L{\'o}pez(2010)}]{Otero:2010}
Pablo~Gamallo Otero and Isaac~Gonz{\'a}lez L{\'o}pez. 2010.
\newblock Wikipedia as multilingual source of comparable corpora.

\bibitem[{Papineni et~al.(2002)Papineni, Roukos, Ward, and Zhu}]{Papineni:2002}
Kishore Papineni, Salim Roukos, Todd Ward, and Wei-Jing Zhu. 2002.
\newblock \href{https://doi.org/10.3115/1073083.1073135}{Bleu: A method for
  automatic evaluation of machine translation}.
\newblock In {\em Proceedings of the 40th Annual Meeting on Association for
  Computational Linguistics\/}. Association for Computational Linguistics,
  Stroudsburg, PA, USA, ACL '02, pages 311--318.
\newblock
  \href{https://doi.org/10.3115/1073083.1073135}{https://doi.org/10.3115/1073083.1073135}.

\bibitem[{Pascanu et~al.(2013)Pascanu, Mikolov, and Bengio}]{Pascanu:2013}
Razvan Pascanu, Tomas Mikolov, and Yoshua Bengio. 2013.
\newblock \href{http://arxiv.org/abs/1211.5063}{On the difficulty of training
  recurrent neural networks}.
\newblock In {\em Proceedings of the 30th International Conference on
  International Conference on Machine Learning - Volume 28\/}. JMLR.org,
  ICML'13, pages III--1310--III--1318.
\newblock
  \href{http://arxiv.org/abs/1211.5063}{http://arxiv.org/abs/1211.5063}.

\bibitem[{Patry and Langlais(2011)}]{Patry:2011}
Alexandre Patry and Philippe Langlais. 2011.
\newblock \href{http://dl.acm.org/citation.cfm?id=2024236.2024252}{Identifying
  parallel documents from a large bilingual collection of texts: Application to
  parallel article extraction in wikipedia}.
\newblock In {\em Proceedings of the 4th Workshop on Building and Using
  Comparable Corpora: Comparable Corpora and the Web\/}. Association for
  Computational Linguistics, Stroudsburg, PA, USA, BUCC '11, pages 87--95.
\newblock
  \href{http://dl.acm.org/citation.cfm?id=2024236.2024252}{http://dl.acm.org/citation.cfm?id=2024236.2024252}.

\bibitem[{Schuster and Paliwal(1997)}]{Schuster:97}
Mike. Schuster and Kuldip~K. Paliwal. 1997.
\newblock \href{https://doi.org/10.1109/78.650093}{Bidirectional recurrent
  neural networks}.
\newblock {\em Trans. Sig. Proc.\/} 45(11):2673--2681.
\newblock
  \href{https://doi.org/10.1109/78.650093}{https://doi.org/10.1109/78.650093}.

\bibitem[{Smith et~al.(2010)Smith, Quirk, and Toutanova}]{Smith:2010}
Jason~R. Smith, Chris Quirk, and Kristina Toutanova. 2010.
\newblock \href{http://dl.acm.org/citation.cfm?id=1857999.1858062}{Extracting
  parallel sentences from comparable corpora using document level alignment}.
\newblock In {\em Human Language Technologies: The 2010 Annual Conference of
  the North American Chapter of the Association for Computational
  Linguistics\/}. Association for Computational Linguistics, Stroudsburg, PA,
  USA, HLT '10, pages 403--411.
\newblock
  \href{http://dl.acm.org/citation.cfm?id=1857999.1858062}{http://dl.acm.org/citation.cfm?id=1857999.1858062}.

\bibitem[{Sutskever et~al.(2014)Sutskever, Vinyals, and Le}]{Sutskever:2014}
Ilya Sutskever, Oriol Vinyals, and Quoc~V. Le. 2014.
\newblock \href{http://dl.acm.org/citation.cfm?id=2969033.2969173}{Sequence to
  sequence learning with neural networks}.
\newblock In {\em Proceedings of the 27th International Conference on Neural
  Information Processing Systems\/}. MIT Press, Cambridge, MA, USA, NIPS'14,
  pages 3104--3112.
\newblock
  \href{http://dl.acm.org/citation.cfm?id=2969033.2969173}{http://dl.acm.org/citation.cfm?id=2969033.2969173}.

\bibitem[{Uszkoreit et~al.(2010)Uszkoreit, Ponte, Popat, and
  Dubiner}]{Uszkoreit:2010}
Jakob Uszkoreit, Jay~M. Ponte, Ashok~C. Popat, and Moshe Dubiner. 2010.
\newblock \href{http://dl.acm.org/citation.cfm?id=1873781.1873905}{Large scale
  parallel document mining for machine translation}.
\newblock In {\em Proceedings of the 23rd International Conference on
  Computational Linguistics\/}. Association for Computational Linguistics,
  Stroudsburg, PA, USA, COLING '10, pages 1101--1109.
\newblock
  \href{http://dl.acm.org/citation.cfm?id=1873781.1873905}{http://dl.acm.org/citation.cfm?id=1873781.1873905}.

\bibitem[{Vinyals et~al.(2014)Vinyals, Toshev, Bengio, and
  Erhan}]{Vinyals:2014}
Oriol Vinyals, Alexander Toshev, Samy Bengio, and Dumitru Erhan. 2014.
\newblock \href{http://arxiv.org/abs/1411.4555}{Show and tell: {A} neural image
  caption generator}.
\newblock {\em CoRR\/} abs/1411.4555.
\newblock
  \href{http://arxiv.org/abs/1411.4555}{http://arxiv.org/abs/1411.4555}.

\bibitem[{Vogel et~al.(1996)Vogel, Ney, and Tillmann}]{Vogel:1996}
Stephan Vogel, Hermann Ney, and Christoph Tillmann. 1996.
\newblock \href{https://doi.org/10.3115/993268.993313}{Hmm-based word alignment
  in statistical translation}.
\newblock In {\em Proceedings of the 16th Conference on Computational
  Linguistics - Volume 2\/}. Association for Computational Linguistics,
  Stroudsburg, PA, USA, COLING '96, pages 836--841.
\newblock
  \href{https://doi.org/10.3115/993268.993313}{https://doi.org/10.3115/993268.993313}.

\end{thebibliography}
\bibliographystyle{acl2017}

\end{document}